\documentclass[letterpaper, 10 pt, conference]{ieeeconf}  

\IEEEoverridecommandlockouts                              

\usepackage{acro}
\usepackage{graphicx}

\usepackage{amsmath}
\usepackage{amssymb}
\usepackage{bm}
\usepackage{siunitx}
\usepackage{makecell}
\usepackage{multirow}
\usepackage{multicol}
\usepackage{caption} 
\usepackage{subcaption}
\usepackage{subcaption}
\usepackage{booktabs} 
\usepackage{tikz}
\usetikzlibrary{decorations.pathreplacing}
\usetikzlibrary{shapes.geometric, arrows, positioning, calc, decorations.pathreplacing}

\usepackage{textcmds}

\usepackage{graphicx}   
\usepackage{tabularx}   
\usepackage{array}
\usepackage{xcolor}     
\usepackage[table]{xcolor}
\usepackage{pgfplots}
\usepackage{pgfplotstable}
\pgfplotsset{compat=1.18}
\usepackage{filecontents}
\usepackage{cite}

\usepackage[symbol]{footmisc}

\graphicspath{ {./images/} }

\usepackage{textcomp}
\usepackage[hidelinks]{hyperref}
\hypersetup{
	pdftitle={Behavior Centric Extraction of Scenarios from Highway Traffic and their Domain-Knowledge-Guided Clustering using CVQ-VAE},
	pdfdisplaydoctitle={true},
	pdfauthor={Niklas Roßberg and Sinan Hasirlioglu and Mohamed Essayed Bouzouraa and Wolfgang Utschick and Michael Botsch},
	pdfsubject={Behavior Centric Extraction and Clustering of Scenarios},
	pdfkeywords={Highway scenario, Validation, Scenarios, Clustering, CVQ-VAE, Scenario-as-Specification, Quantitative Validation},%
}
 
\newcommand\copyrighttext{
	\footnotesize \textcopyright 2026 IEEE. Personal use of this material is permitted. Permission from IEEE must be obtained for all other uses, in any current or future media, including reprinting/republishing this material for advertising or promotional purposes, creating new collective works, for resale or redistribution to servers or lists, or reuse of any copyrighted component of this work in other works. DOI: t.\,b.\,d.}
\newcommand\copyrightnotice{
    \begin{tikzpicture}[remember picture,overlay]
	\node[anchor=south,yshift=10pt] at (current page.south) {\fbox{\parbox{\dimexpr\textwidth-\fboxsep-\fboxrule\relax}{\copyrighttext}}};
	\end{tikzpicture}%
}

\makeatletter
\newcommand\notsotiny{\@setfontsize\notsotiny\@vipt\@viipt}
\newcommand\almostscriptsize{\@setfontsize\almostscriptsize{8.3}{8.3}}
\makeatother
\makeatletter
\IEEEtriggercmd{\reset@font\normalfont\almostscriptsize}
\makeatother
\IEEEtriggeratref{1}

\title{\LARGE \bf
Behavior-Centric Extraction of Scenarios from Highway Traffic Data and their Domain-Knowledge-Guided Clustering using CVQ-VAE  

}

\author{Niklas Roßberg $^{1}$, Sinan Hasirlioglu $^{2}$, Mohamed Essayed Bouzouraa$^{2}$, Wolfgang Utschick$^{3}$ and Michael Botsch$^{1}$
\thanks{$^{1}$Technische Hochschule Ingolstadt, 85049 Ingolstadt, Germany
        {\tt\small \{niklas.rossberg, michael.botsch\}@thi.de}}%
\thanks{$^{2}$AUDI AG, 85045 Ingolstadt, Germany
        {\tt\small \{sinan.hasirlioglu, essayed.bouzouraa\}@audi.de}}
\thanks{$^{3}$Technische Universität München, 80333 München, Germany
        {\tt\small utschick@tum.de}}%
}

\DeclareAcronym{ADS}{
short = ADS ,
long = Automated Driving Systems ,
short-plural = s ,
long-plural = s 
}

\DeclareAcronym{ADF}{
short = ADF ,
long = Autonomous Driving Function ,
short-plural = s ,
long-plural = s 
}

\DeclareAcronym{CCP}{
short = CCP ,
long = Coupon Collector's Problem ,
short-plural = s ,
long-plural = s 
}

\DeclareAcronym{VQ}{
short = VQ,
long = Vector Quantization,
short-plural = s ,
long-plural = s 
}

\DeclareAcronym{CVQ-VAE}{
short = CVQ-VAE ,
long = Clustering Vector Quantized - Variational Autoencoder ,
short-plural = s ,
long-plural = s 
}

\DeclareAcronym{kl}{
short = kl ,
long = keep lane ,
short-plural = s ,
long-plural = s 
}

\DeclareAcronym{lcl}{
short = lcl ,
long = lane change left ,
short-plural = s ,
long-plural = s 
}

\DeclareAcronym{lcr}{
short = lcr ,
long = lane change right ,
short-plural = s ,
long-plural = s 
}

\DeclareAcronym{ML}{
short = ML ,
long = Machine Learning ,
short-plural = s ,
long-plural = s 
}

\DeclareAcronym{DTW}{
short = DTW ,
long = Dynamic Time Warping ,
short-plural = s ,
long-plural = s 
}

\DeclareAcronym{ODD}{
short = ODD ,
long = Operational Design Domain ,
short-plural = s ,
long-plural = s 
}

\DeclareAcronym{OOD}{
short = OOD ,
long = Out-of-Distribution ,
short-plural = s ,
long-plural = s 
}

\DeclareAcronym{PDF}{
short = PDF ,
long = probability density function ,
short-plural = s ,
long-plural = s 
}

\DeclareAcronym{SaS}{
short = SaS ,
long = Scenario as Specification ,
short-plural = s ,
long-plural = s 
}

\DeclareAcronym{VAE}{
short = VAE ,
long = Variational Autoencoder ,
short-plural = s ,
long-plural = s 
}

\DeclareAcronym{VQ-VAE}{
short = VQ-VAE ,
long = Vector Quantized - Variational Autoencoder ,
short-plural = s ,
long-plural = s 
}

\DeclareAcronym{SFM}{
short = SFM ,
long = Social Force Model ,
short-plural = s ,
long-plural = s 
}

\DeclareAcronym{DG-SFM}{
short = DG-SFM ,
long = Directed Gradient - Social Force Model ,
short-plural = s ,
long-plural = s 
}

\DeclareAcronym{THW}{
short = THW ,
long = Time Head Way ,
short-plural = s ,
long-plural = s 
}
\begin{document}

\maketitle
\copyrightnotice
\thispagestyle{empty}
\pagestyle{empty}

\begin{abstract}

Approval of \ac{ADS} depends on evaluating its behavior within representative real-world traffic scenarios. 
A common way to obtain such scenarios is to extract them from real-world data recordings.
These can then be grouped and serve as basis on which the \ac{ADS} is subsequently tested.
This poses two central challenges: how scenarios are extracted and how they are grouped. 
Existing extraction methods rely on heterogeneous definitions, hindering scenario comparability. 
For the grouping of scenarios, rule-based or \ac{ML}-based methods can be utilized.
However, while modern ML-based approaches can handle the complexity of traffic scenarios, unlike rule-based approaches,  they lack interpretability and may not align with domain-knowledge.
This work contributes to a standardized scenario extraction based on the Scenario-as-Specification concept, as well as a domain-knowledge-guided scenario clustering process. 
Experiments on the highD dataset demonstrate that scenarios can be extracted reliably and that domain-knowledge can be effectively integrated into the clustering process.
As a result, the proposed methodology supports a more standardized process for deriving scenario categories from highway data recordings and thus enables a more efficient validation process of automated vehicles.

\end{abstract}


\section{Introduction}
One of the key challenges in developing \ac{ADS} is ensuring safe and reliable operation.
To verify this, scenarios are collected from the system's intended \ac{ODD}.
These scenarios represent the diverse traffic situations an \ac{ADS} may encounter.
The system can then be tested within these scenarios, either on proving grounds or in simulation.
This approach offers the advantage that testing can be significantly accelerated and made more cost-effective compared to real-world testing~\cite{KALRA2016182}.
Consequently, scenario-based testing has become a central paradigm in the research on \ac{ADS} development and validation~\cite{Langner2019, Neurohr2020, Bouzouraa2025}.

However, the foundation for such scenarios is raw traffic recordings, such as the highD dataset~\cite{Krajewski2018}.
The first challenge, therefore, lies in extracting individual scenarios from these recordings~\cite{Neurohr2020}.
A straightforward approach is to define the entire trajectory of a recorded vehicle as one scenario, or to segment it into consecutive, fixed-length snippets~\cite{Epple2020}.
Another approach is to detect events and extract individual scenarios each time an event occurs. 
\begin{figure}
    \centering
    \includegraphics[width=\linewidth]{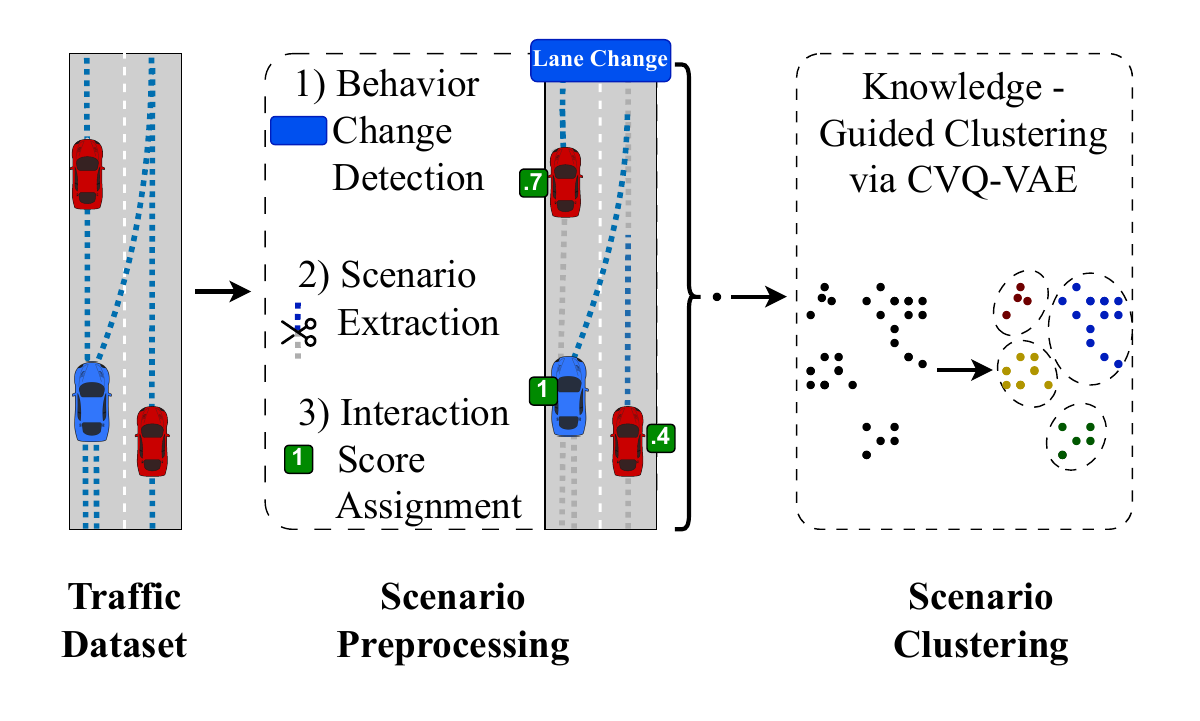}
    \vspace{-0.5cm}
    \caption{Overview of the proposed method for knowledge guided scenario extraction and clustering.}
    \vspace{-0.7cm}
    \label{fig:Overview}
\end{figure}
A wide range of studies determine such events by applying rule-based approaches.
Examples include changes in the surrounding environment \cite{Langner2019, Schuldes2024}, changes in the ego behavior such as lane-change maneuvers \cite{Kerber2020, Neumeier2024}, or violations of a minimum time headway \cite{Balasubramanian2021}.
However, these approaches lack standardization, which can result in scenarios derived by different methods being non-comparable, hindering the development of \ac{ADS}.
To overcome this limitation, Bouzouraa and Hasirlioglu~\cite{Bouzouraa2025} propose the \ac{SaS} approach.
This concept, explained further in section \ref{subsec:Prel-Scenario}, defines the start of a scenario by a change in the ego vehicle’s behavior.
However, detection is nontrivial given heterogeneous behavior changes and ambiguous onsets: lane-change durations vary, measurements are noisy, and maneuvers may be aborted.
Therefore, a threshold-adaptive rule-based method to detect changes in the ego vehicle’s behavior is developed and compared against other methods.
Based on these detected changes, scenarios are subsequently extracted.

Once scenarios are available, the next challenge is grouping them into meaningful categories for testing and analysis as shown in Fig. \ref{fig:Overview}.
Since these categories are not known a priori, this problem can be addressed using unsupervised machine learning.
Scenarios may differ substantially in structure (e.g., number of vehicles or velocities) yet represent the same underlying maneuver, such as a cut-in.
Standard clustering approaches may overemphasize superficial properties and miss the behavioral essence.
Therefore, unsupervised clustering must be guided by domain-knowledge that captures what truly defines scenario similarity.
This work addresses that challenge with an autoencoder-based methodology for domain-knowledge-guided clustering. 
A \ac{CVQ-VAE}~\cite{Zheng2023} is employed, and its latent representation is enriched with domain priors from behavior change detection.
Taking expert-knowledge, such as physical constrains \cite{Egolf2025} or interaction between traffic participants into account when processing scenarios can lead to remarkable results as shown in \cite{Elter2022, Elter2025}. 
Therefore, each vehicle is assigned an interaction score derived from the \ac{DG-SFM}~\cite{Baden2025}, which directs the model toward the influential agents and ensures that clusters reflect the underlying interaction logic rather than raw scenario complexity. 
The key contributions of this paper are as follows:

\begin{enumerate}
\item A standardized rule-based methodology for highway scenario extraction based on ego vehicle behavior changes.

\item A domain-knowledge-guided scenario clustering approach that injects knowledge about vehicle interactions and ego behaviors into a \ac{CVQ-VAE}.

\item An evaluation of behavior change detection and scenario clustering using ground-truth datasets, demonstrating the effectiveness of the methods.

\end{enumerate}


\section{Related Work}
To enable scenario-based validation of \ac{ADS}, continuous traffic recordings must be segmented into comparable units and collected in a scenario library. 
Section~\ref{subsec:RW-Extraction} reviews different methods to segment traffic recordings and subsequently extract scenarios. Section~\ref{subsec:RW-Clustering} surveys approaches to cluster the extracted scenarios.

\subsection{Extraction of Traffic Scenarios}
\label{subsec:RW-Extraction}
For the extraction of scenarios rule-based methods remain prevalent since they are simple, interpretable, and easy to scale. 
They typically define scenario boundaries via explicit triggers  on ego-centric signals~\cite{Kerber2020, Balasubramanian2021, Balasubramanian2023} or interactions \cite{Chang2024, King2021}, and specify termination either by additional state conditions or by a fixed duration. 
Single-feature triggers use thresholds on ego measures, e.g., a scenario starts when the time headway to the leading vehicle falls below a limit, with a fixed end time~\cite{Balasubramanian2021}.
Similarly, predefined scenario types with fixed durations are instantiated once ego conditions such as “decelerating” and “leader present” hold~\cite{Balasubramanian2023}. 
Event-specific rules detect maneuvers directly, e.g., lane changes from lane-ids and lateral-velocity thresholds to set start/end around the lane crossing~\cite{Kerber2020}. 
Tag-based pipelines generalise this idea by composing semantic tags over ego state, relative kinematics, and environment context to extract segments~\cite{Guo2023, Gelder2020,  Hartjen2019}.
Interaction-centered rules trigger scenarios when ego and neighbor trajectories merge, diverge, or cross \cite{King2021}.
Recent work~\cite{Chang2024} refines the interaction-centered scenarios through improved interaction and relevance metrics to prioritise influential neighbors. 
While effective, these approaches can be sensitive to thresholds and struggle to generalise across diverse scenario types~\cite{Cai2022}.








On the other hand, \ac{ML}-based methods learn scenario structure directly from data and can therefore complement rule-based pipelines~\cite{Erdogan2019, Montanari2021}. 
Broadly, they fall into two groups: (i) Latent-clustering approaches, which encode time steps, cluster the embeddings, and place boundaries where cluster assignments change~\cite{Kreutz2022, Hernández2022, Chetouane2022}.
(ii) Probabilistic labeling approaches, which predict per-timestep probabilities for predefined scenario classes and derive segments from consecutive high-probability windows~\cite{Montanari2021, Elspas2021}. 

Within the first group, Kreutz et al.~\cite{Kreutz2022} use k-means on self-supervised embeddings to detect latent regime shifts, while Chetouane et al.~\cite{Chetouane2022} compare alternative clustering algorithms for episode extraction. 
As supervised baselines, Elspas et al.~\cite{Elspas2021} train fully convolutional networks on programmatically generated labels to obtain per-timestep probabilities.
Montanari et al.~\cite{Montanari2021} couple a rule-based state machine with an RNN to aid the  transition detection. 
A hybrid alternative first segments by maximising an energy objective and then classifies the resulting variable-length maneuvers~\cite{Aboah2023}.
Overall, these approaches lessen a-priori tuning and handle variable durations. 
Nevertheless, as demonstrated by~\cite{Erdogan2019}, even simple rule-based baselines achieve competitive accuracy.

\subsection{Clustering of Traffic Scenarios}
\label{subsec:RW-Clustering}
The methodologies for clustering traffic scenarios found in literature can roughly be grouped into two different approaches: (i) rule-based similarities \cite{Guo2023, Kerber2020, Gelder2020, Ries2021} and (ii) learning-based clustering \cite{Neumeier2024, Balasubramanian2023, Kruber2019, Zhao2021, Wurst2022, Hauer2020, Zipfl2023, Zeng2025}.

Within rule-based methods, distance-driven approaches compare trajectories directly: Kerber et al.~\cite{Kerber2020} align vehicle positions over a scenario, Ries et al.~\cite{Ries2021} pre-filter by present object types and maneuvers and then measure similarity via \ac{DTW} on feature distances.

Tag-based pipelines  categorize scenarios using semantic tags from extraction (ego state, relative kinematics, context)~\cite{Gelder2020, Guo2023}.

Learning-based approaches replace hand-crafted metrics with embeddings and data-adaptive similarities. Random-Forest approaches derive proximities from tree paths: Kruber et al.~\cite{Kruber2019} train with synthetic noise to capture real-data structure and cluster via path proximity. 
Balasubramanian et al.~\cite{Balasubramanian2023} extend this approach with open-set/open-world scenario discovery. 
Hauer et al.~\cite{Hauer2020} combine \ac{DTW} with k-means on latent features to group similar scenarios. Other methods embed scene graphs with contrastive learning before clustering \cite{Zipfl2023}.
Zeng et al.~\cite{Zeng2025} introduce Toeplitz Inverse Covariance Clustering (TICC) for segmenting short windows into stable action clusters.
Autoencoder-based approach including the \ac{VQ-VAE} \cite{Oord2017} and metric-guided variants have gained traction for their ability to compress multi-agent spatiotemporal structure into compact, clustering-friendly representations \cite{Neumeier2024, Wurst2022, Zhao2021, Rossberg2025, Zheng2023, Fertig2024}. 
However, the integration of domain-knowledge into the clustering process of scenarios remains an open challenge.
To the best of our knowledge, no existing work combines rule-based scenario extraction together with explicit guidance from that extraction to drive the scenario clustering.

\section{Preliminaries}
This section introduces important terms and the methods used and extended in this work. 
Section~\ref{subsec:Prel-Scenario} outlines the scenario concept and its structure.
Section~\ref{subsec:Prel-DG-SFM} shows a method utilized to enrich scenarios with domain-knowledge.

\subsection{Scenario Definition and Structure}
\label{subsec:Prel-Scenario}
The \ac{SaS} approach treats scenarios as formal, testable specifications of the system~\cite{Bouzouraa2025}, enabling their systematic use throughout development and for approval. 
It defines a scenario based on the target behavior of the ego vehicle.
Target behavior means the specified correct behavior for the \ac{ADS} in the current scenario.
A new scenario arises only when the target behavior changes.
In other words: Changes in the static or dynamic environment constitute a new scenario only if they imply a different target behavior.
\emph{Example 1:} The ego travels on a highway in the same lane, following a lead vehicle. Although external conditions may vary (overtakes, straight or curved road), it remains the same scenario as long as the target behavior “keep lane and maintain car-following” is unchanged.
\emph{Example~2:}~A cut-in forces the ego to decelerate to re-establish a safe time headway. The target behavior changes to “decelerate due to cut-in until a safe distance is restored,” and a new scenario is instantiated. For further details of the \ac{SaS} approach, see \cite{Bouzouraa2025}.

Real-world recordings cannot provide the target behavior. 
However, it is essential to test against real traffic scenarios~\cite{Guo2023}.
To overcome this problem, scenarios are detected via observed behavior changes of the ego, referred to as ego behavior changes.
This ensures comparability across scenarios from different sources. 
Since the focus lies on dynamic motion the \textit{highD} dataset is selected as traffic dataset~\cite{Krajewski2018}.

The temporal extent of each scenario is constrained to a fixed observation horizon $T_{\text{obs}}$, following each detected ego behavior change.
This event-anchored trimming aligns samples in time, and thereby enables stable training with a shared codebook in the \ac{CVQ-VAE}.
The horizon $T_{obs}$ is chosen to capture the behavior change and the subsequent steady behavior.
Also, the spatial cardinality is chosen to be fixed. 
A constant number $N$ of vehicles per scenario is enforced. 
If fewer than $N$ are present, pseudo-vehicles, which do not influence the clustering, are added until $N$ is reached. 
Each trajectory is represented by $F$ features per vehicle, yielding a fixed-size tensor  for each scenario $\bm{\xi}$:
\[
\bm{\xi} \in \mathbb{R}^{N \times F \times T_{\text{obs}}}.
\]
Concrete parameter choices are given in Section~\ref{sec:Methodology}.
To inject domain-knowledge, a rule-based pseudo-class $\bm{s} \in \mathbb{R}^{S}$ encodes the ego’s behavior change type (Section~\ref{subsec:BCDandSE}). 
In addition, an interaction score matrix $\bm{T} \in \mathbb{R}^{N \times T_{obs}}$ is computed per vehicle and time step using the \ac{DG-SFM}~\cite{Baden2025} to quantify each neighbor’s relevance to the ego maneuver. 
The resulting dataset is
\[
\mathcal{D}=\bigl\{ \bigl(\bm{\xi}^{(m)},\, \bm{s}^{(m)},\, \bm{T}^{(m)} \bigr) \bigr\}_{m=1}^{M},
\]
comprising $M$ samples, each containing the scenario tensor~$\bm{\xi}$, its pseudo-class~$\bm{s}$, and interaction scores~$\bm{T}$.

\subsection{Directed Gradient - Social Force Model}
\label{subsec:Prel-DG-SFM}
The \ac{DG-SFM} \cite{Baden2025} adapts the repulsive potential component of the Social Force Model (SFM) \cite{helbing1995} from crowd dynamics to road traffic.
The authors use it to align a Transformer's attention over scenario participants with a human-interpretable, physics-based interaction score. 
The SFM represents interactions between agents through repulsive potentials, i.e., virtual forces that discourage agents from coming too close to each other. 
In road traffic, such potentials quantify how strongly one vehicle affects another’s motion and therefore serve as an interpretable proxy for interaction~\cite{Baden2025}.
DG-SFM modifies this repulsive potential by shaping it asymmetrically, stretching the influence area forward in the agent’s direction of motion, which results in an “egg-shaped” interaction field~\cite{Baden2025}.

Let $i$ denote the ego vehicle, and let $J_i$ be the set of its neighboring vehicles. 
For each \mbox{$\ell\in{i,j}$}, let \mbox{$r_\ell\in\mathbb{R}^2$} and \mbox{$v_\ell\in\mathbb{R}^2$} denote the position and velocity of vehicle $\ell$, respectively.
DG-SFM defines two complementary components: (i) how deeply $j$ intrudes into $i$'s directional personal space ($\hat\beta^{A}_{ij}$), and (ii) the short-horizon change of the interaction based on $j$'s potential field ($\hat\beta^{B}_{ij}$):
\begin{align}
\hat\beta^{A}_{ij} &= V_{\mathrm{egg}}(r_j,\,r_i,\,v_i),\\
\hat\beta^{B}_{ij} &= V_{\mathrm{egg}}(r_i^{*},\,r_j^{*},\,v_j)\;-\;V_{\mathrm{egg}}(r_i,\,r_j,\,v_j),
\end{align}
where the short-horizon extrapolation is given by \mbox{$r_\ell^{*}=r_\ell+N_{\!DG}\,v_\ell\,\Delta t$} \mbox{($\ell\!\in\!\{i,j\}$)}, with $\Delta t>0$ denoting the temporal resolution and $N_{DG}$ the number of discretization steps \cite{Baden2025}.
$V_{\mathrm{egg}}$ is the direction-aware repulsive potential along the mover's heading (see \cite{Baden2025}).
The components are subsequently weighted and normalized to yield a deterministic interaction distribution over all neighbors.

\section{Methodology}
\label{sec:Methodology}
This section presents our approach to scenario extraction and the integration of domain-knowledge into the subsequent clustering stage. The \ac{CVQ-VAE} serves as the foundation for clustering the scenarios. The prior knowledge incorporated originates from two sources: (i) behavior change detection and (ii) the \ac{DG-SFM}.

\subsection{Problem Description}

\label{sec:problem}

We consider continuous multi-agent highway recordings sampled at $\Delta t$.
Event-anchored, fixed-size scenarios must be extracted from this. 
Each scenario is represented as \mbox{$\bm{\xi}\in\mathbb{R}^{N\times F\times T_{\text{obs}}}$},
where $N = 9$, $F = 6$ $(x,y,v_x,v_y,a_x,a_y)$, and $T_{\text{obs}} = 4 \si{s} $. 
Each scenario is further enriched with two sources of prior knowledge:
\mbox{(i) a} rule-based pseudo-class $\bm{s}\in\mathbb{R}^{S}$ that one-hot encodes the ego’s behavior change type, and
(ii) an interaction score matrix $\bm{T}\in[0,1]^{N\times T_{\text{obs}}}$ that quantifies per-agent relevance for each timestep, derived from the \ac{DG-SFM}~\cite{Baden2025}.

The objective is to learn a representation
\begin{equation}
    \bm{\hat{z}} = h_\theta(\bm{\xi}) \in \mathbb{R}^d
\end{equation}

that captures both the ego’s behavior and the interaction structure with surrounding vehicles.
By incorporating the prior knowledge $(\bm{s},\bm{T})$ obtained during scenario preprocessing, the learned representation is guided to reflect domain-relevant semantics, enabling data-driven yet interpretable clustering into traffic scenario categories \mbox{$q \in \{1,\dots,Q\}$}.

\subsection{Behavior Change Detection and Scenario Extraction}
\label{subsec:BCDandSE}
Scenario extraction based on ego vehicle behavior changes first requires identifying such changes.
Therefore, each car in the dataset is treated once as the ego vehicle.
Discrete trajectories sampled at a constant time step~$\Delta t$, with available longitudinal and lateral accelerations~$(a_x(t),a_y(t))$ and lateral velocity~$v_y(t)$ are assumed. 
From these signals, sequences of consistent motion states are derived using adaptive thresholding rules that distinguish between different modes of longitudinal and lateral behavior.

\textbf{Longitudinal behavior:}  
Longitudinal motion is classified using an adaptive threshold applied to the acceleration $a_x(t)$. 
The key idea is to ignore short, transient fluctuations and to identify only sustained deviations from previous behavior as genuine behavior changes.
A transition from the \emph{zero} state (coasting) to the \emph{acceleration}  state is triggered once
\[
|a_x(t)| > \tau_{\mathrm{up}} \quad \text{for at least } n_{\mathrm{up}} \text{ consecutive frames}.
\]
Multiple threshold--duration pairs $(\tau_{\mathrm{up}},n_{\mathrm{up}})$ are applied in parallel, e.g., $(0.2,100),(0.3,50),(0.4,25)$, in order to detect both mild but persistent accelerations and shorter but stronger bursts.
Analogously, the transition back to the \emph{zero} state occurs when $|a_x(t)|<\tau_{\mathrm{down}}$ for $n_{\mathrm{down}}$ frames. 
If the acceleration magnitude exceeds a high threshold $\tau_{\mathrm{extreme}}$, the state immediately switches to \emph{extreme}, capturing emergency-like maneuvers. 
The same applies to \emph{deceleration}, but with negative thresholds.

\textbf{Lateral behavior:}  
Lateral maneuvers are detected by accumulating the lateral displacement
\begin{equation}
\Delta y = \sum_{t=t_0}^{t_1} v_y(t)\,\Delta t
\end{equation}
over intervals where the sign of $v_y(t)$ remains constant. 
This ensures that only consistent lateral movements are considered. 
If the displacement magnitude exceeds a lane-change threshold $\tau_{\mathrm{LC}}$, the interval is classified as a \emph{lane change}; otherwise, it is labeled as \emph{keep lane}. 
This criterion captures completed lane changes while filtering out small oscillations within the lane. 

\textbf{Post-processing:}  
After the initial labeling, consecutive frames with the same state are combined into segments, and short segments $(n < 3)$ are removed to improve robustness. 
Each segment is then assigned a composite label $\ell(t)$ that integrates both longitudinal behavior (e.g., \emph{zero}, \emph{normal}, \emph{extreme}) and lateral behavior (e.g., \emph{keep lane}, \emph{lane change}). 
A behavior change occurs whenever this composite label changes. 
Formally, valid transition points $t_c$ are defined as
\begin{equation}
\mathcal{C}=\{\,t_c \mid \ell(t_c-\Delta t)\neq \ell(t_c+\Delta t)\,\},
\end{equation}
which mark the onset of new behaviors. $\mathcal{C}$ denotes the set of all detected change points $t_c$. Consecutive lane-change segments are further merged into a single composite segment, with their associated longitudinal acceleration state attached. 

\textbf{Scenario extraction:}  
Each detected change point serves as the anchor for a fixed-length temporal window \mbox{$[t_c-50,\,t_c+75]$}, provided that sufficient trajectory data is available. 
This yields ego-event-centered scenarios that are temporally aligned to behavioral changes. 
For the entire time window, the following features are extracted for the ego vehicle and all surrounding vehicles: position $(x,y)$, velocity $(v_x,v_y)$, and acceleration $(a_x,a_y)$.
The pseudo-class label $s$ is constructed from the behavior states before and after each behavior point $t_c$ and stored as a one-hot encoded vector. The total number of distinct pseudo-classes is denoted by $S$.
\begin{figure*}
    \centering
    \includegraphics[width=\linewidth]{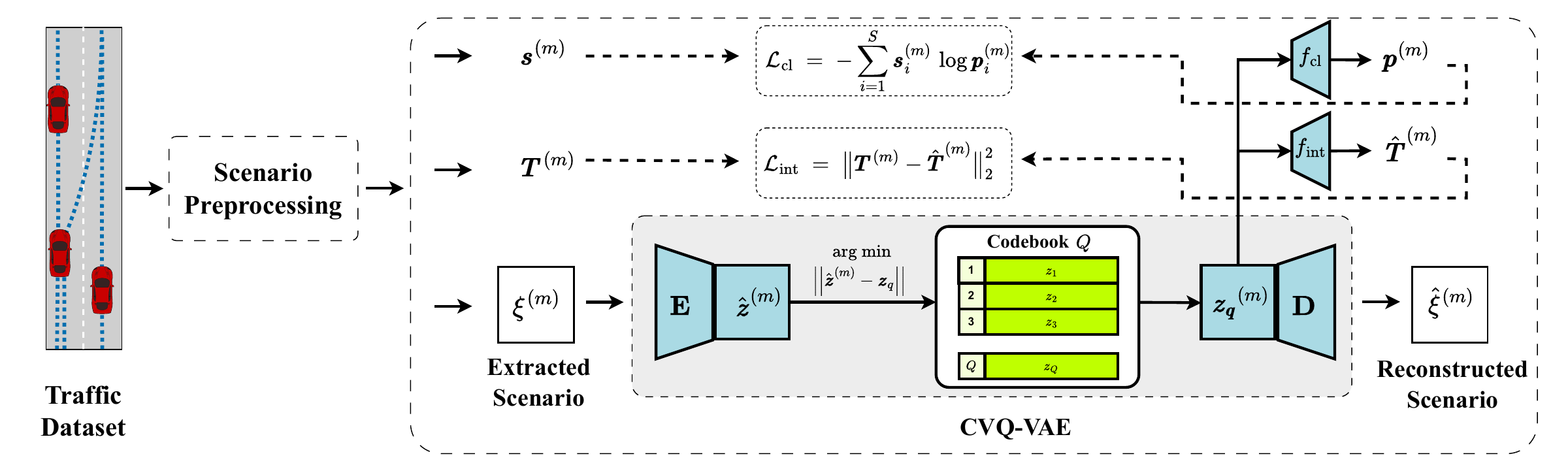}
    \caption{\small In the Scenario Preprocessing stage, ego–behavior changes are detected and used to extract scenarios from the traffic dataset. In addition, the interaction matrix $\bm{T}^{(m)}$ and the pseudo-class label vector $\bm{s}^{(m)}$ are computed for each scenario. For clustering, a \ac{CVQ-VAE} with a predefined number of codebook entries $Q$ is employed. The model receives only the scenario trajectories $\bm{\xi}^{(m)}$ as input, produces a discretized representation $\bm{z_q}^{(m)}$, and predicts both the interaction matrix and the behavior class from this latent representation. }
    \vspace{-0.5cm}
    \label{fig:Pipeline}
\end{figure*}

\subsection{Enriching Scenarios with Domain-Knowledge}

Beyond the pseudo-class label $s$, each scenario is enriched with per\mbox{-}frame interaction scores that quantify how strongly each neighboring vehicle influences the ego behavior.
For each time step~$t$ and each neighbor~\mbox{$j\in J_i(t)$} of ego $i$, the directional repulsive components $\hat\beta^{A}_{ij}(t)$ and $\hat\beta^{B}_{ij}(t)$ are computed~\cite{Baden2025}.
These components are combined into a single interaction score
\begin{equation}
  \beta_{ij}(t)
  \;=\;
  \tau_{\mathrm{sum}}\,\hat\beta^{A}_{ij}(t)
  \;+\;
  \bigl(1-\tau_{\mathrm{sum}}\bigr)\,\hat\beta^{B}_{ij}(t),
\end{equation}
with $\tau_{\mathrm{sum}}\in[0,1]$. Scores are then normalized framewise over the present neighbors using a softmax, yielding normalized interaction scores $\pi_{ij}(t)$ for each neighbor $j$ at time step~$t$~\cite{Baden2025}.
The normalized scores are assembled into the interaction matrix
\[
  \bm{T}\in[0,1]^{N\times T_{\text{obs}}},
  \qquad
  \bm{T} \;=\; \bigl[\,\bm{t}(1)\; \cdots\; \bm{t}(T_{\text{obs}})\,\bigr],
\]
where $\bm{t}(t)\in[0,1]^N$ denotes the column at time $t$ with components
\[
  \bm{t}_n(t)=
  \begin{cases}
    1, & \text{if } n \text{ is the ego row},\\[4pt]
    \pi_{ij}(t), & \text{if } n \text{ corresponds to neighbor } j\in J_i(t),\\[4pt]
    0, & \text{otherwise (absent slot)}.
  \end{cases}
\]
If fewer than $N$ vehicles are present, missing slots are padded with zeros to preserve shape.
The ego row is always set to $1$ for all $t$, thereby marking the identity of the ego.

\subsection{Model Architecture for Traffic Scenario Clustering} 
\label{subsec:Met-Scenario-Clustering}
For the clustering step, the \ac{CVQ-VAE} is employed. 
The \ac{CVQ-VAE} provides a stable backbone, onto which domain-knowledge is integrated.
In contrast to conventional autoencoders with continuous latents, vector quantization discretizes the latent space into a finite set of codebook entries as shown in Fig. \ref{fig:Pipeline}, each of which can be interpreted as a cluster. 
This property makes the model particularly attractive for structured domains such as traffic scenarios, where discrete categories and stable cluster assignments are essential.
Similar VQ-based approaches have demonstrated their effectiveness for categorical representation learning \cite{Oord2017, Neumeier2024, Rossberg2025}.
Here, the mechanism is leveraged to ensure a reliable partitioning of scenarios into a predefined number of clusters~$Q$, while the proposed extensions ensure that these clusters are semantically meaningful and consistent with domain reasoning.

Given a scenario  $\bm{\xi}^{(m)}\in\mathbb{R}^{N\times F\times T_{\text{obs}}}$, the encoder~$h_\theta$ maps the input to a continuous latent
$\hat{\bm{z}}^{(m)} = h_\theta \bigl(\bm{\xi}^{(m)}\bigr)$ with \mbox{$\hat{\bm{z}}^{(m)} \in \mathbb{R}^{d}$}.
Instead of passing $\hat{\bm{z}}^{(m)}$ to the decoder, it is discretized by nearest-neighbor lookup in a finite codebook 
\mbox{$\mathcal{Z}=\{\bm{z}_1,\dots,\bm{z}_Q\}$} with $\bm{z}_q\in\mathbb{R}^{d}$:
\begin{equation}
\label{eq:argminvq}
\bm{z}^{(m)}_{q} \;=\; \arg\min_{\bm{z}\in\mathcal{Z}} \bigl\|\,\hat{\bm{z}}^{(m)} - \bm{z}\,\bigr\|_2^{2}, 
\qquad q\in\{1,\dots,Q\}.
\end{equation}
The decoder $g_\psi$ then reconstructs the input as \mbox{$\hat{\bm{\xi}}^{(m)} = g_\psi\!\bigl(\bm{z}^{(m)}_{q}\bigr)\in\mathbb{R}^{N\times F\times T_{\text{obs}}}$}.
By construction, vector quantization maps each scenario to one of the $Q$ codebook entries, thereby inducing a clustering with a predefined number of traffic scenario categories~$Q$. Further details are given in \cite{Rossberg2025}.

\paragraph{CVQ-VAE Loss}
The baseline objective combines reconstruction, vector-quantization, and commitment terms:
\begin{equation}
\begin{aligned}
\mathcal{L}_{\text{cvq}}
&= \bigl\|\bm{\xi}^{(m)}-\hat{\bm{\xi}}^{(m)}\bigr\|_2^2
 + \bigl\|\mathrm{sg}[\hat{\bm{z}}^{(m)}]-\bm{z}_q^{(m)}\bigr\|_2^2 \\
&\qquad \qquad \qquad \qquad + \bigl\|\hat{\bm{z}}^{(m)}-\mathrm{sg}[\bm{z}_q^{(m)}]\bigr\|_2^2.
\end{aligned}
\end{equation}
where $\mathrm{sg}[\cdot]$ denotes the stop-gradient operator.

\paragraph{Knowledge-guided Losses}
To inject domain-knowledge, the discretized latent $\bm{z}_q^{(m)}$ drives two linear prediction heads that do not feed into the decoder but act as auxiliary supervision signals for clustering:

\textbf{Pseudo-class head:}
As depicted in Fig. \ref{fig:Pipeline}, a linear classifier $f_{\text{cl}}(\cdot)=$ is added to the \ac{CVQ-VAE}. 
It  maps the discretized latent scenario representation $\bm{z}_q^{(m)}$ to logits in~$\mathbb{R}^{S}$:
\begin{equation}
\bm{p}^{(m)} \;=\; \mathrm{softmax}\!\bigl(f_{\text{cl}}(\bm{z}_q^{(m)})\bigr), 
\quad \text{with } f_{\text{cl}}:\mathbb{R}^d \to \mathbb{R}^{S}.
\end{equation}
The classifier is trained to predict the one-hot pseudo-class label \mbox{$\bm{s}^{(m)}\in{0,1}^{S}$} that is constructed during scenario extraction.
Therefore, a cross-entropy loss is employed, which directly measures the discrepancy between the predicted class distribution and the expert-defined pseudo-class.
This pushed the latent representation $\bm{z}_q^{(m)}$ to linearly separable regions that correspond to the intended ego maneuver categories:
\begin{equation}
\label{eq:lcl}
\mathcal{L}_{\text{cl}}
\;=\; -\sum_{i=1}^{S} \bm{s}^{(m)}_{i}\,\log \bm{p}^{(m)}_{i}.
\end{equation}
This objective teaches the model to preserve principal ego maneuver information in the latent vector $\bm{z_q}$ and to assign each scenario to clusters with similar ego maneuvers, thereby discouraging clusters that mix heterogeneous ego behaviors.

\textbf{Interaction head:}
Additionally, a linear projection head \mbox{$f_{\text{int}}$}, also shown in Fig. \ref{fig:Pipeline}, maps $\bm{z}_q^{(m)}$ into an interaction matrix prediction $\hat{\bm{T}}^{(m)}\in[0,1]^{N\times T_{\text{obs}}}$,
\begin{equation}
\hat{\bm{T}}^{(m)} \;=\; \sigma\!\bigl(f_{\text{int}}(\bm{z}^{(m)}_{q})\bigr),
\quad \text{with } f_{\text{int}}:\mathbb{R}^d \to \mathbb{R}^{N\times T_{\text{obs}}}.
\end{equation}
where $\sigma(\cdot)$ denotes an elementwise sigmoid to ensure $[0,1]$ range and the reshape aligns with the target shape.
A squared error penalizes deviations:
\begin{equation}
\label{eq:lint}
\mathcal{L}_{\text{int}} \;=\; \bigl\| \bm{T}^{(m)} - \hat{\bm{T}}^{(m)} \bigr\|_2^2.
\end{equation}
Intuitively, the interaction head aligns clusters with a coherent spatiotemporal interaction pattern across agents.
In doing so, the model is steered to recognize characteristic interaction structures and to discount incidental, low-relevance vehicles when forming clusters.
Together with the pseudo-class head, this discourages mixing scenarios that differ in behavioral logic, even when their kinematics are superficially similar.

\paragraph{Total objective}
The overall training loss is the weighted sum
\begin{equation}
\label{eq:ltotal}
\mathcal{L}_{\text{total}}
\;=\;
\mathcal{L}_{\text{cvq}}
\;+\;
\lambda_{\text{cl}}\,\mathcal{L}_{\text{cl}}
\;+\;
\lambda_{\text{int}}\,\mathcal{L}_{\text{int}},
\end{equation}
with $\lambda_{\text{cl}},\lambda_{\text{int}}\!\ge\!0$ controlling the strength of knowledge guidance.
This converts clustering from a purely data-driven grouping into a knowledge-guided process. Traffic scenario clusters become discrete, interpretable, and aligned with domain semantics captured by $(\bm{s},\bm{T})$.

\section{Evaluation}

This section describes the dataset used in the experiments and outlines the implementation details.
It further evaluates the performance of the behavior change detection and examines the impact of integrating domain-knowledge into the clustering process.

\subsection{Dataset}

In this work, the publicly available highD dataset \cite{Krajewski2018} is used. 
It comprises highway trajectory recordings at 25 \si{Hz} collected at six locations in Germany. 
To ensure comparability across scenarios, the analysis is restricted to recordings with three lanes. 
For the clustering part in this study, we focus on categories in which the ego transitions from keep lane to lane change \mbox{(KL $\rightarrow$ LC)}, thereby reducing the problem scope. 
The resulting selection contains $16,768$ scenarios. 
Of these, 85\% are used for training and 15\% are held out as a validation split during training.

\subsection{Behavior Change Detection}

For validation, $100$ vehicles and their trajectories from the \emph{highD} dataset were manually annotated.
Behavior changes were labeled whenever a lane change was observable or a sustained change in speed occurred.
Because the exact onset is not always unambiguous, each event was represented by a $50$-frame $(= 2\,\si{s})$ window during which the change takes place.
In total 119 behavior changes were detected.
Composite labels were assigned as described in Sec.~\ref{subsec:BCDandSE}.
A predicted change point $\hat t_c$ is counted as a true positive if it falls inside the corresponding ground-truth window and the composite label matches, otherwise it is counted as a false positive.
Ground-truth events without a matching prediction contribute to false negatives.
Precision, recall, and the counts (TP/FP/FN) are reported.

\begin{table}[t]
\centering
\small
\vspace{0.3 cm}
\caption{Evaluation of behavior--change detection on the manually annotated subset. Best values are in \textbf{bold}.}
\label{tab:bcd_validation}
\begin{tabular}{lcccccc}
\toprule
Method & Precision $\uparrow$ & Recall $\uparrow$ & TP & FP & FN \\
\midrule
Rule-based & \textbf{0.741} & \textbf{0.916} & 109 & 38  & 10 \\
EMA \cite{Aboah2023} & 0.196 & 0.244 & 29  & 119 & 90 \\
CVQ-VAE & 0.302 & 0.723 & 86  & 199 & 33 \\
\bottomrule
\end{tabular}
\vspace{-0.5 cm}
\end{table}

To benchmark the introduced rule–based behavior change, two alternative methods are introduced.
First, the exponential moving average (EMA) change detector proposed in \cite{Aboah2023} is applied.
The EMA segments the time series by dilating a temporal window around each time step $t$, computing the scaled window energy, and declaring events at local energy maxima.
Window sizes of $\{30,60,90\}$ are evaluated to probe short-, mid-, and longer–range smoothing.
As a data–driven alternative, all trajectories from the highD dataset are partitioned into fixed–length snippets and embedded with a separate \mbox{CVQ–VAE}.
The embeddings are assigned to 64 clusters during the unsupervised training.
At test time, ego trajectories are again segmented into snippets and each snippet is assigned to a cluster. 
A behavior change is declared whenever the assigned cluster switches between consecutive snippets of the same ego trajectory. The results for the approaches are presented in table \ref{tab:bcd_validation}.

The results show that the rule–based detection substantially outperforms the two alternative approaches. 
The EMA baseline detects at least one behavior change per trajectory by design, which leads to a large number of false positives.
Moreover, it often fails to align with the relatively gradual behavior changes observed on highways.
While the unsupervised \ac{CVQ-VAE} clustering is conceptually well-suited to capture complex patterns, it achieves poor performance.
The strong imbalance in highway data, where most vehicles exhibit near-constant motion, means that substantially larger and more diverse datasets would be required to detect behavior changes reliably.
In contrast, highway behavior changes follow well-structured and domain-specific traffic dynamics that can be formalized explicitly, enabling the rule-based method to exploit domain-knowledge effectively.
Consequently, mainly for smaller traffic datasets, expert-driven rules remain a reliable option for behavior change detection in this context.

\subsection{Traffic Scenario Clustering}
\label{subsec:scenario_clustering_eval}

\begin{figure}
    \centering
    \includegraphics[width=\linewidth]{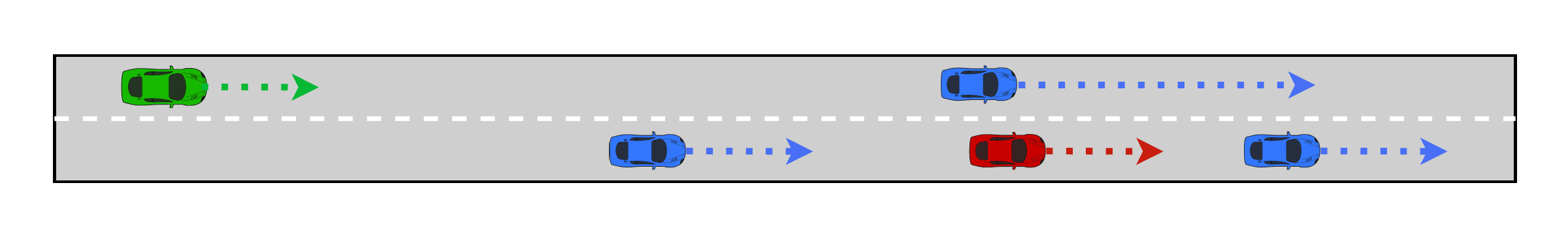}
    \caption{Exemplary augmented scenario where the additional vehicle (green) has no influence on the ego vehicle (red).}
    \vspace{-0.7cm}
    \label{fig:Scen-Aug}
\end{figure}

The previously described subset of data is clustered into \mbox{$Q = 64$} clusters. To evaluate to which extent it is required to introduce domain-knowledge into the clustering process, a random subset of previously extracted scenarios was augmented. 
Specifically, 50 scenarios were generated in which an additional vehicle was inserted at a larger distance from the ego vehicle. 
This inserted vehicle maintains a constant velocity and does not perform a lane change. 
To ensure realistic motion patterns, the trajectory of the added vehicle was taken from the actual trajectory of a suitable vehicle in the dataset. 
An illustration of a single time step $t$ of such an augmented scenario is shown in Fig.~\ref{fig:Scen-Aug}. 
During augmentation, care was taken to ensure that the added vehicle has no influence on the observed ego behavior. 
Accordingly, its interaction was fixed to $t_{\text{aug}} = 0$ for all observation steps $T_{\text{obs}}$.
The underlying idea is to evaluate how well the model captures behaviorally relevant motion patterns.

Two metrics are employed to assess clustering quality and the contribution of domain-knowledge. 
The first metric is based on pseudo-classes that represent the ego behavior. 
For each latent codebook entry $\bm{z}_q$, the linear classifier predicts the probability distribution over all pseudo-classes, as described in Eq.~(9). 
If the model consistently predicts a single pseudo-class for a given cluster, this indicates that the cluster contains scenarios belonging to the same behavioral category. 
Conversely, if predictions are spread across multiple pseudo-classes, this suggests that the cluster mixes different behaviors. 
To quantify this effect, the Shannon entropy \mbox{$H_{\text{avg}} = \mathbb{E}_{q=1...Q}[H_q]$} is computed across all codebook entries \cite{Neumeier2024}.

The entropy $H_q$ for each entry $q$ is defined as
\begin{equation}
    H_q = - \sum_{i=1}^S p_{\text{cl},i} \log_2 p_{\text{cl},i},
\end{equation}
where $p_{\text{cl},i}$ denotes the predicted probability of pseudo-class $S_i$ for the latent representation $\bm{\hat{z}}^{(m)}$. 
The entropy ranges from complete purity \mbox{($H_q = 0$)} to maximum impurity \mbox{($H_q = \log_2(10) \approx 3.322$)} given $S=10$ classes.

The second metric evaluates the consistency of cluster assignments for the augmented scenarios. 
For each augmented variant, we verify whether it is assigned to the same cluster as its corresponding original scenario. 
Since the augmentation was explicitly designed to have no effect on the ego behavior, consistent assignments are expected. 
The fraction of correctly matched pairs over the total number of augmented scenarios defines an accuracy measure:
\begin{equation}
    accuracy = \frac{\text{correct assignments}}{\text{total number of augmented scenarios}}.
\end{equation}

In a first step, the CVQ-VAE was trained on the original scenarios with the knowledge-guided loss weights set to $\lambda_{\text{cl}} = 0$ and $\lambda_{\text{int}} = 0$, i.e., without incorporating domain-knowledge. 
During inference, the augmented scenarios were passed through the model, and their cluster assignments were evaluated. 
Based on these assignments, the two metrics were computed and reported in Table~\ref{tab:eval} under the column “no DK” (no domain-knowledge). 
In the next step, the weighting parameters were set to $\lambda_{\text{cl}} = 1$ and $\lambda_{\text{int}} = 1$, thereby integrating domain-knowledge into the training. 
The corresponding results are listed in Table~\ref{tab:eval} under the column “DK”.

\begin{table}[t]
\centering
\small
\vspace{0.3 cm}
\caption{Clustering on the train split: cluster-purity ($\downarrow$) and augmentation cluster accuracy ($\uparrow$). DK = Domain-Knowledge. Best values per row are in \textbf{bold}.}
\label{tab:eval}
\begin{tabular}{lcccc}
\toprule
\multirow{2}{*}{Backend} & \multicolumn{2}{c}{no DK} & \multicolumn{2}{c}{DK} \\
\cmidrule(lr){2-3}\cmidrule(lr){4-5}
 & purity $\downarrow$ & accuracy $\uparrow$ & purity $\downarrow$ & accuracy $\uparrow$ \\
\midrule
CVQ-VAE      & 3.014 & 0.068 & 1.243 & \textbf{0.568} \\
k-means      & 3.014 & \textbf{0.091} & 1.127 & 0.182 \\
Hierarchical & \textbf{3.010} & 0.068 & \textbf{1.153} & 0.250 \\
\bottomrule
\end{tabular}
\vspace{-0.5 cm}
\end{table}

In addition to the CVQ-VAE codebook clustering, two further clustering approaches were applied in order to consolidate the evaluation.
Specifically, $k$-means clustering and hierarchical agglomerative clustering were performed directly on the latent scenario representations $\hat{\bm{z}}$ obtained from the CVQ-VAE encoder. 
By including these baseline methods, the evaluation allows for a fair assessment of whether improvements are specific to the CVQ-VAE quantization mechanism or can also be observed when applying standard clustering algorithms to the learned latent representations.

The results are summarized in Table~\ref{tab:eval}. 
Without domain-knowledge, all three clustering methods exhibit low augmentation accuracy, indicating that augmented scenarios are frequently assigned to clusters different from their original counterparts. 
In parallel, the high entropy values reveal that multiple ego maneuvers are mixed within single clusters, which demonstrates the difficulty of separating behaviors solely from the raw latent representations. 
Once domain-knowledge is injected, clustering quality improves consistently across all methods: both purity loss decreases and augmentation accuracy increases. 
The effect is most pronounced for CVQ-VAE codebook clustering. 
Here, augmentation accuracy rises from $\!0.068$ to $\!0.568$, demonstrating that directly integrating domain priors through the codebook and auxiliary heads provides stronger guidance than applying generic clustering algorithms to the latent space. 
Although purity is not the best among all methods, the CVQ-VAE nevertheless achieves competitive results. 
This indicates that CVQ-VAE clusters, while slightly less homogeneous, are far more consistent in preserving the behavioral identity of scenarios across augmentations
These findings indicate that the dataset alone does not provide enough variability to support reliable data-driven clustering, making domain-knowledge indispensable for learning meaningful scenario semantics.

\section{Conclusion}

In this work we introduced a rule-based approach for extracting highway scenarios by detecting ego–behavior changes, thereby providing a standardized and interpretable mechanism for deriving scenarios from highway data recordings.
Second, we proposed a domain-knowledge-guided clustering framework that incorporates information about vehicle interactions and ego vehicle behavior changes.
Experiments performed on the highD dataset show that training without such domain-knowledge  produces behaviorally mixed clusters and low assignment consistency.
When domain-knowledge is incorporated, the clusters become more coherent and the assignments more stable.
This demonstrates that, under limited scenario availability, injecting expert knowledge provides a clear advantage for clustering.
The largest gains are achieved with CVQ-VAE codebook clustering. 

A current limitation of this study is the use of fixed-length windows, which can cut off or merge behaviors at segment boundaries.
Future work will therefore focus on clustering variable-length scenarios with a varying number of traffic participants.

\vspace{-0.8em}

\section*{Acknowledgment}
We appreciate the funding of this work by Audi AG.

\bibliographystyle{IEEEtran}
\bibliography{Literature}

\end{document}